\documentclass{article}
\usepackage{ijcai18}

\usepackage{times}
\usepackage{xcolor}
\usepackage{soul}
\usepackage[utf8]{inputenc}
\usepackage[small]{caption}
\usepackage{booktabs} 
\usepackage[ruled,linesnumbered]{algorithm2e}
\usepackage{graphicx}
\usepackage{subcaption}
\usepackage{amsmath}
\usepackage{amssymb}
\usepackage{url}

\title{Whatever Does Not Kill Deep Reinforcement Learning, Makes It Stronger\thanks{This work was supported by the National Science Foundation (NSF) (NSF-CRII-CPS-1743490). Any opinions, findings, and conclusions or recommendations expressed in this material are those of the author and do not necessarily reflect the views of the NSF.}}

\author{
Vahid Behzadan$^1$, 
Arslan Munir$^2$, 
\\ 
$^1$ Department of Computer Science, Kansas State University \\
$^2$ Department of Computer Science, Kansas State University\\
behzadan@ksu.edu,
amunir@ksu.edu
}

\begin{document}

\maketitle

\begin{abstract}
Recent developments have established the vulnerability of deep Reinforcement Learning (RL) to policy manipulation attacks via adversarial perturbations. In this paper, we investigate the robustness and resilience of deep RL to training-time and test-time attacks. Through experimental results, we demonstrate that under noncontiguous training-time attacks, Deep Q-Network (DQN) agents can recover and adapt to the adversarial conditions by reactively adjusting the policy. Our results also show that policies learned under adversarial perturbations are more robust to test-time attacks. Furthermore, we compare the performance of $\epsilon$-greedy and parameter-space noise exploration methods in terms of robustness and resilience against adversarial perturbations.
\end{abstract}

\section{Introduction}
Over the past few years, the field of deep Reinforcement Learning (RL) has enjoyed rapidly growing attention and advancements. An important attraction of deep RL lies in exploiting the well-established feature learning capabilities of deep neural networks in tandem with classical RL, thereby enabling end-to-end policy learning through direct mappings of raw observations to actions. This ability extends the domain of classic RL applications to high-dimensional and complex planning problems, leading to human-level and superhuman performance various settings, ranging from learning to playing the game of Go \cite{silver2016alphago} and Atari games \cite{mnih2015human} to robotic manipulation \cite{gu2017deep} and autonomous navigation of aerial \cite{zhang2016learning} and ground \cite{zhu2017target} vehicles. While the interest in adopting deep RL solutions is expanding into more critical domains, such as intelligent transportation systems \cite{atallah2017next}, finance \cite{deng2017deep} and national infrastructure \cite{mohammadi2017semi}, state of the art in understanding and ensuring the security and reliability of such solutions remains at preliminary stages. 

Recently, Behzadan and Munir \cite{behzadan2017} reported the vulnerability of deep RL to both test-time and training-time adversarial attacks exploiting adversarial examples \cite{goodfellow2014explaining}. While further investigations of this vulnerability (e.g., \cite{huang2017adversarial}, \cite{kos2017delving}) verify the fragility of deep RL agents to adversarial perturbations, only a few (e.g., \cite{lin2017detecting}) focus on mitigation and countermeasures, most of which are based on detection and mitigation of test-time attacks. 

In this work, we aim to further the research on security of deep RL by investigating the resilience and robustness of such algorithms to both training-time and test-time attacks. In this context, resilience refers to the ability of an agent to recover from adversarial attacks, and robustness is the ability of an agent to maintain desired functionality under adversarial perturbations. Building on similar attack methodologies as those proposed in \cite{behzadan2017} and \cite{kos2017delving}, we perform an experimental exploration to study practical attack scenarios in which the attacker perturbs training-time observations in a noncontiguous manner - that is, instead of perturbing all observations, each observed state is perturbed with a probability $P(attack)$. Inspired by the proposals of \cite{mandlekar2017adversarially}, we experimentally study the resilience of DQN agents training in Atari 2600 game environments under such attacks. We also investigate the robustness of policies trained under noncontiguous attacks to test-time perturbations. Furthermore, we comparatively study the impact of exploration methods on resilience and robustness of DQN agents. In particular, we focus on the newly proposed parameter-space noise technique. Recent proposals by OpenAI \cite{plappert2017parameter} and DeepMind \cite{fortunato2017noisy} demonstrate that the addition of adaptive noise to the parameters of deep RL architectures may enhance the exploration behavior and convergence speed of such agents. Contrary to classical exploration heuristics such as $\epsilon$-greedy \cite{sutton1998reinforcement}, parameter-space noise is iteratively and adaptively applied to the parameters of the learning model, such as weights of the neural network. Considering the reported enhancements enabled by parameter-space noise exploration, our experiments aim to compare the performance of this technique and $\epsilon$-greedy under adversarial perturbations.

Accordingly, the main contributions of this paper are as follows:

\begin{enumerate}
	\item We demonstrate that Deep Q-Network (DQN) agents are capable of recovering from noncontiguous training-time attacks by adjusting their policies to account for adversarial perturbations.
	\item We demonstrate that the policies trained under adversarial perturbations are more robust to test-time attacks.
	\item We investigate the effect of exploration methods on resilience and robustness by comparing the performance of $\epsilon-greedy$ and parameter-space noise techniques  under adversarial attacks.
\end{enumerate}

The remainder of this paper is organized as follows: Section \ref{background} reviews the relevant background on DQN, parameter-space noise exploration via the NoisyNet approach, and adversarial example attacks on deep RL. Section \ref {Model} describes the training and test-time attack models considered in this work. Section \ref{Setup} details the experiment setup and presents the corresponding results. Section \ref{Conclusion} concludes the paper with remarks on the obtained results and notes on potential venues of further research.

\section{Background}\label{background}
\subsection{Q-Learning}

The problem of RL -i.e. learning to act well through experiential interactions with the environment, is often formally modeled as a Markov Decision Process (MDP). Such processes are described by the tuple $MDP = (S, A, P, R)$, where $S$ is the set of reachable states in the process, $A$ is the set of allowed actions, $P$ represents the transition probabilities (i.e., dynamics), and $R$ is the mapping of transitions to the immediate reward. At any given time step $t$, the environment is at a state $s_t\in S$. The RL agent's choice of action at time $t$, $a_t \in A$ triggers a transition from $s_t$ to a state $s_{t+1}$ according to the transition probability $P_{s_t , s_{t+a}}^{a_t}$. The agent receives a reward $r_t = R(s_t, a_t) \in \mathbb{R}$, where $\mathbb{R}$ denotes the set of real numbers, representing a scalar reward due to the transition from state $s_t$ to state $s_{t+1}$. 

The mechanism for choosing actions at given states is captured in a policy $\pi$. When this mechanism is deterministic, the policy $\pi: S\rightarrow A$ is a mapping between the states and their corresponding desired actions. Similarly, a stochastic policy is a probability distribution over actions that determines the probability of taking each action at any state $s$. The core objective of RL is to learn the optimal policy $\pi^\ast$ that maximizes the cumulative reward over time at time $t$, denoted by the return function $\hat{R}_t = \sum_{k = 0}^{\infty} \gamma^{k} r_{t+k}$, where $\gamma \in [0,1]$ is the discount factor accounting for the diminishing worth of rewards obtained further in time, which also ensures that $\hat{R}$ is bounded.

One approach to this problem is Q-learning, which aims to estimate the value of taking each action in any state, defined as the sum of future rewards when taking that action and following the optimal policy thereafter. Accordingly, the value of an action $a$ in a state $s$ is given by the action-value function $Q$ defined as:
\begin{eqnarray} \label{bellman}
Q(s,a) = R(s, a) + \gamma max_{a'}(Q(s',a'))
\end{eqnarray}

Where $s'$ is the state that emerges as a result of action $a$, and $a'$ is an allowed action for state $s'$. The optimal policy is hence given by $\pi^\ast(s) = \arg\max_a Q(s,a)$

The Q-learning method estimates the optimal action policies by using the Bellman equation $Q_{i+1} (s,a) = \mathbf{E}[R + \gamma \max_a Q_i]$ as the iterative update of a value iteration technique. Practical implementation of Q-learning is commonly based on function approximation of the parametrized Q-function $Q(s,a; \theta) \approx Q^\ast (s,a)$. A common technique for approximating the parametrized non-linear Q-function is to train a neural network whose weights correspond to the parameter vector $\theta$. Such neural networks, commonly referred to as Q-networks, are trained with the goal of minimizing the loss function:

\begin{eqnarray}
L_i(\theta_i) = \mathbf{E}_{s, a\sim \rho(.)} [(y_i - Q(s,a,;\theta_i))^2]
\end{eqnarray}

where $y_i = \mathbf{E}[R + \gamma \max_{a'}Q(s',a';\theta_{i-1}) | s,a]$, and $\rho(s,a)$ is a probability distribution over states $s$ and actions $a$. This optimization problem is typically solved using computationally efficient techniques such as Stochastic Gradient Descent (SGD) \cite{baird1999gradient}.
%
%
%
%
%
%
%

\subsection{Deep Q Networks}

%
%
%
%

Classical Q-networks present a number of shortcomings in the Q-learning process. First, the sequential processing of consecutive observations breaks the \emph{iid} (Independent and Identically Distributed) requirement of training data as successive samples are correlated. Furthermore, slight changes to Q-values leads to rapid changes in the policy estimated by Q-network, which leads to rapid oscillations in $Q$ estimations. Also, since the scale of rewards and Q values are unknown, the gradients of Q-networks can become very large and the rewards can become unbounded.

A deep Q network (DQN) \cite{mnih2015human} is a multi-layered Q-network designed to mitigate such disadvantages. To overcome the issue of correlation between consecutive observations, DQN employs a technique named \emph{experience replay}: Instead of training on successive observations, experience replay samples a random batch of previous observations stored in the replay memory to train on. As a result, the correlation between successive training samples is broken and the iid setting is re-established. In order to avoid oscillations, DQN fixes the parameters of the optimization target $y_i$. These parameters are then updated at regular intervals by adopting the current weights of the Q-network. The issue of unbounded rewards is also solved in DQN by normalizing the reward values to the range $[-1,+1]$, thus preventing Q values from becoming too large.

Mnih et al. \cite{mnih2015human} demonstrate the application of this new Q-network technique to end-to-end learning of Q values in playing Atari games based on raw pixel values observed in the game environment. To capture time-dependent features such as movements, Mnih et al. stack 4 consecutive frames of the game together as the input to the network. For training the Q-network, a random batch is sampled from the previous observation tuples $(s_t, a_t, r_t, s_{t+1})$. In the original proposal of Atari DQN, each observation is processed by 2 layers of convolutional neural networks to learn the features of input frames, which are then employed by feed-forward layers to approximate the Q-function. The target network $\hat{Q}$, with parameters $\theta^{-}$, is synchronized with the parameters of the original $Q$ network at fixed intervals -i.e., at every $i$th iteration,  $\theta^-_{t} = \theta_t$, and is kept fixed until the next synchronization. The target value for optimization of DQN learning thus becomes:

\begin{eqnarray}
y'_t \equiv r_{t+1} + \gamma max_{a'} \hat{Q}(S_{t+1}, a'; \theta^-)
\end{eqnarray}

Accordingly, the training objective can be stated as:

\begin{eqnarray}\label{SGD}
min_{a_t} (y'_t - Q(s_t, a_t, \theta))^2
\end{eqnarray}

\subsection{NoisyNets}
Introduced by Fortunato et al. \cite{fortunato2017noisy}, NoisyNet is a type of neural network whose biases and weights are iteratively perturbed during training by a parametric function of the noise. Such a neural network can be represented by $y = f_\theta (x)$, parametrized by the vector of noisy parameters $\theta = \mu +\Sigma \ast \epsilon$, where $\tau = (\mu, \Sigma)$ is a set of vectors representing learnable parameters, $\epsilon$ is a vector of zero-mean noise with fixed statistics, and $\ast$ is element-wise multiplication. Accordingly, \cite{fortunato2017noisy} proposes a modified DQN algorithm. This algorithm omits $\epsilon$-greedy, and instead optimizes the Q-function greedily. The fully connected layers of the Q-function are parametrized as a NoisyNet, whose parameters are drawn from the noisy network parameter distribution after every replay step. The noise distribution proposed for this architecture is factorized Gaussian noise. At experience replay, current NoisyNet parameter samples are held constant, while during the optimization of each action step, these parameters are re-sampled. The loss function in this variant of DQN is minimizing. The parametrized action-value function $Q(x,a,\epsilon; \tau)$ can be seen as a random variable, and accounted for accordingly in the optimization function. Further technical details of this architecture are available in \cite{fortunato2017noisy}, with a similar proposal detailed in \cite{plappert2017parameter}. 

\subsection{Adversarial Examples}
In \cite{szegedy2013intriguing}, Szegedy et al. report that several machine learning models, including deep neural networks, are vulnerable to misclassify inputs that are only slightly different from correctly classified samples drawn from the data distribution. Such inputs are known as adversarial examples. Furthermore, it was shown that a wide variety of models with different architectures trained on sampled subsets of the training data misclassify the same adversarial example.

This problem can be formalized as follows: consider a machine learning system $M$ and a benign input sample $C$ which is correctly classified by the machine learning system, i.e. $M(C) = y_{true}$. According to Szegedy \cite{szegedy2013intriguing} and many proceeding studies \cite{papernot2016limitations}, it is possible to construct an adversarial example $A = C + \delta$, which is perceptually indistinguishable from $C$, but is classified incorrectly, i.e. $M(A) \neq y_{true}$.

Adversarial examples are misclassified far more often than examples that have been perturbed by random noise, even if the magnitude of the noise is much larger than the magnitude of the adversarial perturbation \cite{goodfellow2014explaining}. According to the objective of adversaries, adversarial example attacks are generally classified into the following two categories: \emph{misclassification} (non-targeted) attacks, which aim for generating examples that are classified incorrectly by the target network, and \emph{targeted} attacks, whose goal is to generate samples that the target misclassifies into an arbitrary class designated by the attacker.
	
To generate such adversarial examples, several algorithms have been proposed, such as the Fast Gradient Sign Method (FGSM) by Goodfellow et al., \cite{goodfellow2014explaining}. A limiting assumption in many of the crafting algorithms is that the attacker has complete knowledge of the target neural network, including its architecture, weights, and other hyperparameters. To overcome this constrain, Papernot et al. \cite{papernot2016practical} proposed a blackbox approach to generating adversarial examples. This method exploits the generalized nature of adversarial examples: An adversarial example crafted for a neural network classifier applies to most other neural network classifiers that perform the same classification task, regardless of their architecture and hyperparameters. Accordingly, the approach proposed in \cite{papernot2016practical} is based on generating a replica of the target model. The attacker creates and trains this model over a dataset from a mixture of samples obtained by observing target's response to test inputs, as well as synthetically generated inputs and output pairs. Once trained, any of the adversarial example crafting algorithms that require knowledge of the target network can be applied to the replica. Due to the transferability of adversarial examples, the perturbed inputs generated for the replica model will induce misclassifications in many of the other networks that perform the same task. 

\subsection{Adversarial Example Attacks on Deep RL}
Building on the fact that classification and estimation of Q-network are both instances of function approximation, Behzadan and Munir \cite{behzadan2017} propose that adversarial examples also apply to policy learning in Deep RL. Their paper demonstrates that adversarial examples can manipulate the choice of actions at test-time, and disrupt the policy learning process via continuous perturbation of observations at training-time. While most extensions of this work focus on test-time attacks (e.g., \cite{huang2017adversarial}, \cite{lin2017detecting}), Kos and Song \cite{kos2017delving} build on these results to show that training-time disruption of policy can also be achieved with noncontiguous attacks: In practical settings, it is unlikely for the adversary to successfully perturb all observations. Therefore, \cite{kos2017delving} investigate the scenario in which each observation is perturbed with a probability $P(attack)$, and experimentally analyze various heuristics for optimizing the impact of such attacks on policy disruption. In the following section, we adopt this idea to develop our attack model and experimental settings. 

\section{Attack Model}\label{Model}
We consider an attacker whose goal is to perturb the optimality of actions taken by a DQN agent at either the test-time or training-time by perturbing the observations of the agent. Aiming for the analysis of worst possible cases, we consider whitebox attack scenarios, i.e., the adversary is assumed to have complete knowledge of the target model's architecture and hyperparameters, as well as state observations. In this attack model, the attacker has no direct influence on the target's architecture and parameters, including its reward function, exploration rate, parameter noise, and the optimization mechanism. The only parameter that the attacker can directly manipulate is the configuration of the training environment observed by the target. For instance, in the case of Atari games \cite{mnih2015human}, the attacker may manipulate pixel values of the game frames, but not the score. We assume that the adversary acts as a Man-In-The-Middle (MITM) between its target and the environment, capable of manipulating the state before it is observed by the target through either predicting future states, having a quicker action speed than the target's sampling rate, or inducing a delay between generation of the new environment and its observation by the target. We also assume that the adversary perturbs each observation of the targeted agent with the probability $P(attack) \in [0,1]$. To avoid detection and minimize influence on the environment's dynamics, we impose an extra constraint on the attack such that the magnitude of perturbations applied to each configuration must be smaller than a set value denoted by $\epsilon$. Also, we do not limit the attacker's domain of perturbations, e.g., the attacker may perturb any of the pixel values in any frame of an Atari game environment. Perturbations are crafted using the FGSM technique for non-targeted attacks \cite{goodfellow2014explaining}, which aims to find an adversarial perturbation that causes the target to output any action other than the correct one. It is noteworthy that in this attack model, the adversary does not require any knowledge of the target's reward function, environmental dynamics, or exploration techniques, and only needs to know the weights of target's Q-network at any given time-step.
 
In training-time attacks, the adversary manipulates target DQN's learning process by crafting states $s'_t = s_t + \delta_{s_t}$ such that $\max_{a}{\hat{Q}(s'_{t}, a; \theta^{-}_{t})}$ identifies an incorrect choice of optimal action. If the attacker is capable of crafting adversarial inputs $s'_t$ such that the value of equation \ref{SGD} is minimized for any action $a'_t \neq a_t$, then the policy learned by DQN is skewed towards suggesting $a'$ as the optimal action given the state $s'_t$, which at the feature learning level of DQN, can be indistinguishable from the unperturbed $s_t$ if $\delta_{s_t}$ is sufficiently small.  

For test-time attacks, a similar MITM attack model is considered. To analyze the robustness of policies under worst possible conditions, the adversary applies non-targeted FGSM perturbations to all observations ($P(attack) = 1$).

%
%

\section{Experimental Results} \label{Setup}
Following the classic benchmarks of DQN, our experimental environments consist of emulated Atari 2600 games. We demonstrate the results for two game environments, namely Breakout and Pong. To enable the comparison between $\epsilon$-greedy and NoisyNet exploration techniques, we run our training and test-time experiments on both implementations. The neural network configuration of both models, as well as the $\epsilon$-greedy algorithm follow that of the original DQN proposal by Mnih et al. \cite{mnih2015human}, and the NoisyNet implementation is based on the algorithm presented in \cite{fortunato2017noisy}.

We implemented the experimentation platform in TensorFlow using OpenAI Gym \cite{brockman2016openai} for emulating the game environment and Cleverhans \cite{papernot2016cleverhans} for crafting the adversarial examples. Our DQN implementation is a modified version of the module in OpenAI Baselines \cite{baselines}. The source-code for our platform is available at \cite{RLAttack} for open-source use in further research.

\subsection{Training-time Attacks}\label{TrainingAttacks}
In accordance to the attack model detailed in Section \ref{Model}, we investigate the performance of DQN with both $\epsilon$-greedy and NoisyNet exploration techniques under noncontiguous attacks. Similar to the work in \cite{kos2017delving}, our experiments initiate training-time attacks at or close to the convergence of mean return towards the optimal value. Maximum perturbation threshold for FGSM is set to 0.004.

\begin{figure*}[t]
	\begin{subfigure}[h]{0.45\textwidth}
		\includegraphics[width=\textwidth]{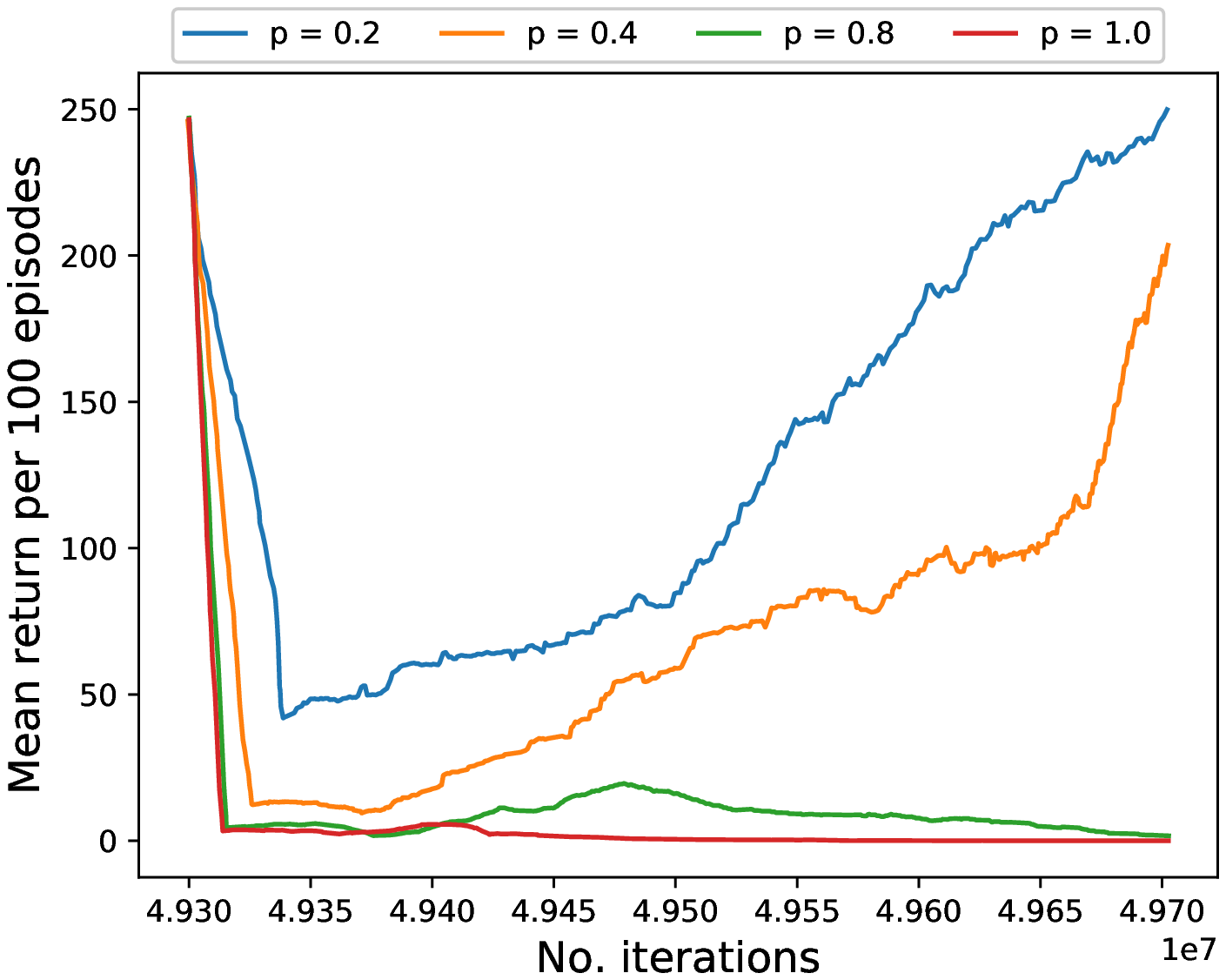}
		\caption{NoisyNet}
	\end{subfigure}
	\begin{subfigure}[h]{0.45\textwidth}
		\includegraphics[width=\textwidth]{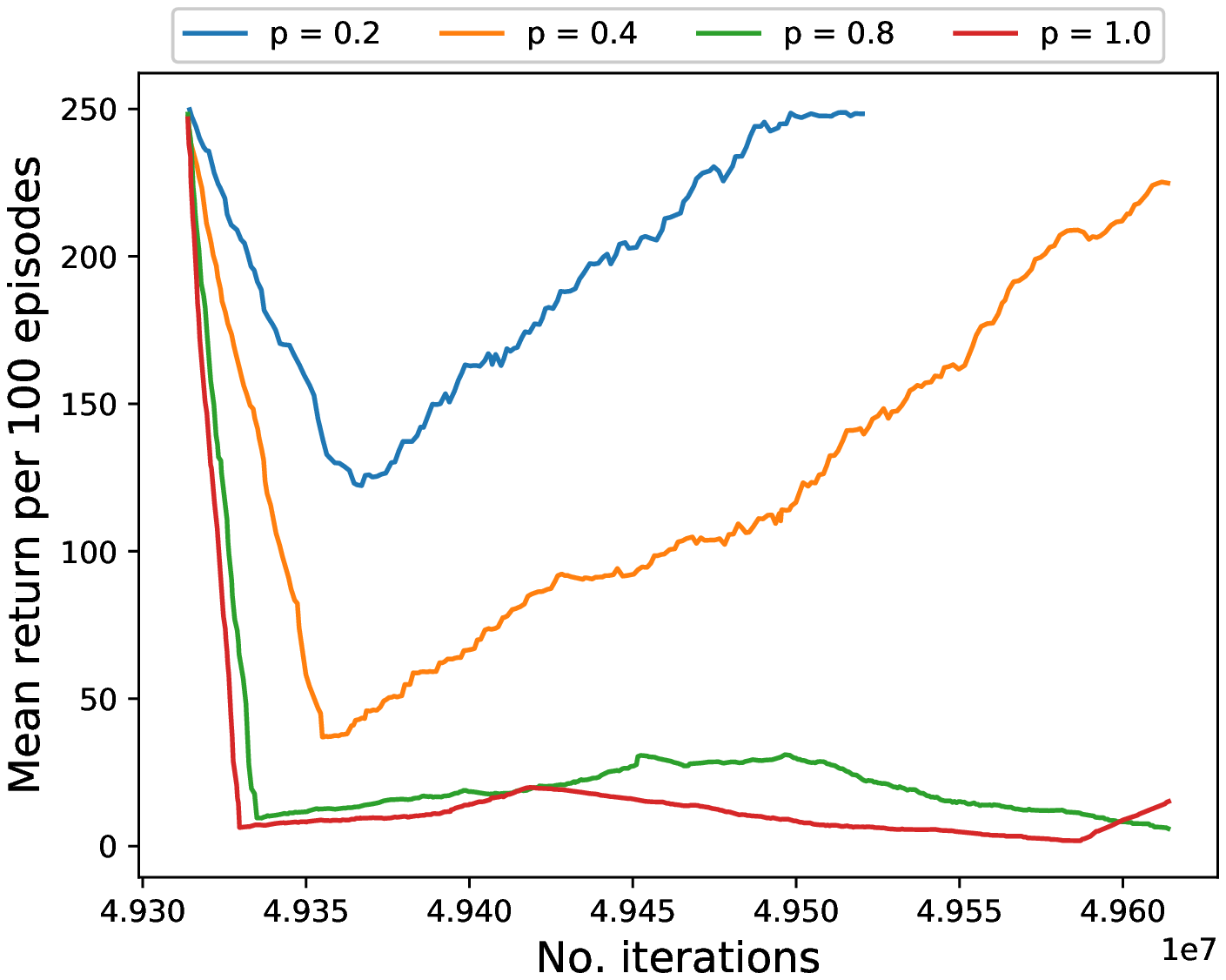}
		\caption{$\epsilon$-greedy}
	\end{subfigure}
	\caption{Breakout - training-time attacks ($p = P(attack)$)} 
	\label{break-train}
\end{figure*}

Figure \ref{break-train} illustrates the results of noncontiguous attacks in Breakout for varying values of $p = P(attack)$. It can be seen that in both NoisyNet and $\epsilon$-greedy implementations, for $p = 0.2$ and $p = 0.4$, agents recover from policy disruption and adjust their policies to overcome adversarial perturbations. Conversely, for $p = 0.8$ and $p = 1.0$ (contiguous attack), both agents fail to recover from policy disruptions in the span of our experiments.

\begin{figure*}[t]
	\begin{subfigure}[h]{0.45\textwidth}
		\includegraphics[width=\textwidth]{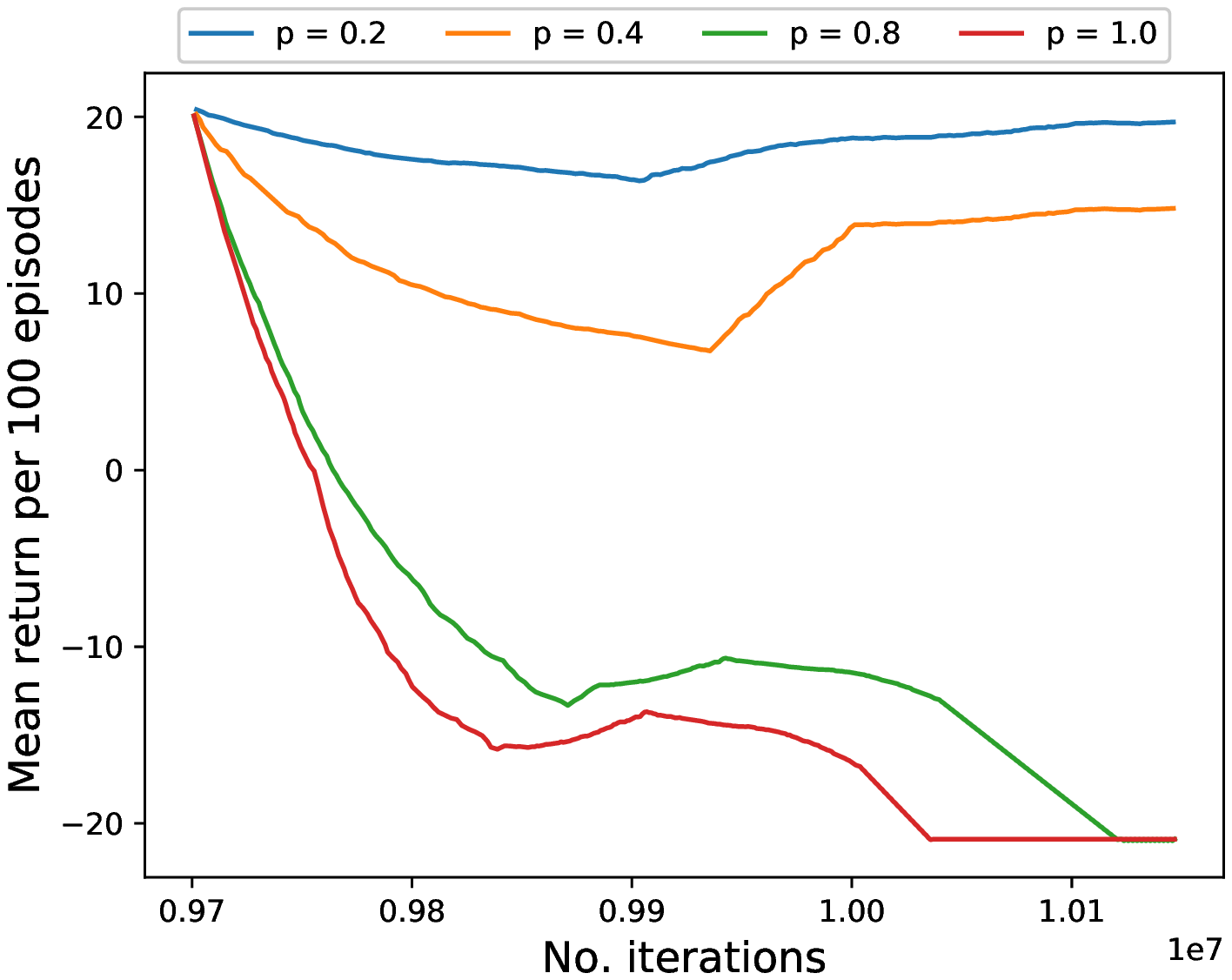}
		\caption{NoisyNet}
	\end{subfigure}
	\begin{subfigure}[h]{0.45\textwidth}
		\includegraphics[width=\textwidth]{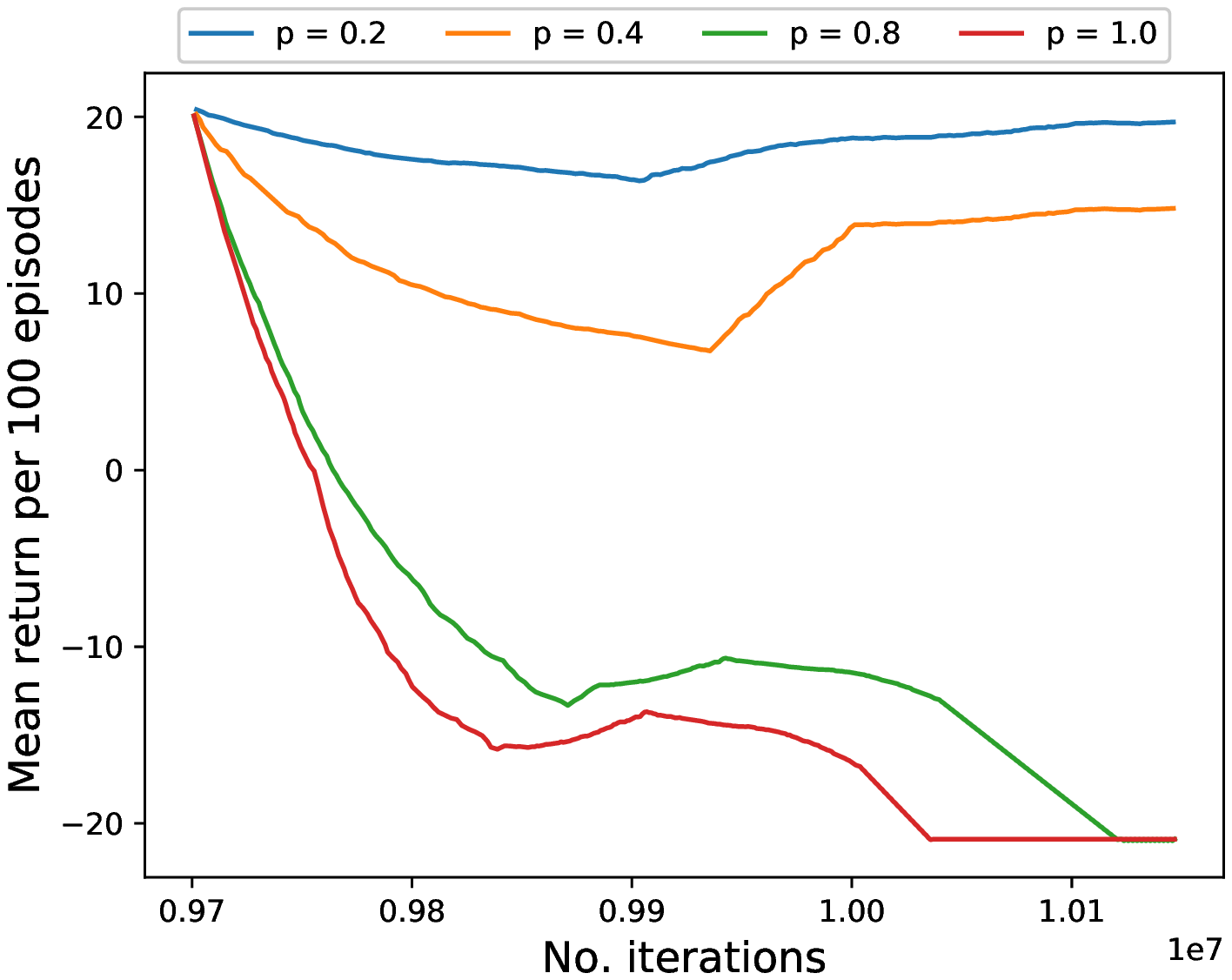}
		\caption{$\epsilon$-greedy}
	\end{subfigure}
	\caption{Pong - training-time attacks ($p = P(attack)$)} 
	\label{pong-train}
\end{figure*}

Figure \ref{pong-train} depicts a similar behavior for the Pong environment in a wider span over iterations. Both sets of results verify that if the attack probability is less than some threshold, DQN agents are able to recover from training-time adversarial attacks. An interesting observation is the existence of phase transition points: in both Pong and Breakout and with both exploration mechanisms, for $p = 0.2$ and $p = 0.4$, the training performance deteriorates almost uniformly until a minimum point is reached, from which onward the agent begins to recover and adjust the policy towards optimal performance. An intuitive interpretation of this behavior stems from the statistics of experience replay: for the agent to recover from adversarial perturbations, the number of interactions with the perturbed environment must reach a critical mass (i.e., minimum) so that the randomly sampled batches from the experience buffer can represent the statistical significance of perturbations. 

Furthermore, it can be seen that in both cases for $p = 0.2$ and $p = 0.4$, the minimum value of return (averaged over most recent 100 episodes) reached by $\epsilon$-greedy agents is higher than that of NoisyNet agents, which may be attributed to better robustness of $\epsilon$-greedy over NoisyNet. Conversely, NoisyNet agents reach phase transition and recovery in fewer iterations than their $\epsilon$-greedy counterparts, which can be evidence of superior resilience in NoisyNet over $\epsilon$-greedy agents.

\begin{figure*} [h]
	\centering
	\begin{subfigure}[!h]{0.45\textwidth}
		\includegraphics[width=\textwidth]{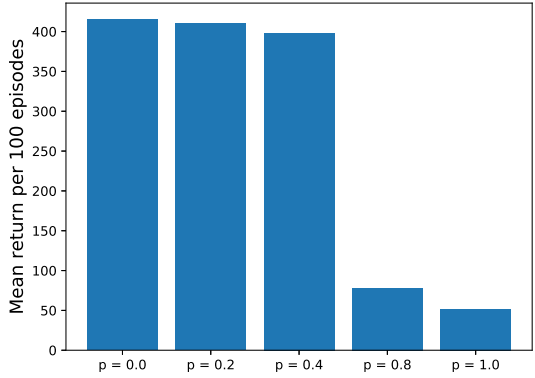}
		\caption{Breakout-NoisyNet}
	\end{subfigure}
	\begin{subfigure}[!h]{0.45\textwidth}
		\includegraphics[width=\textwidth]{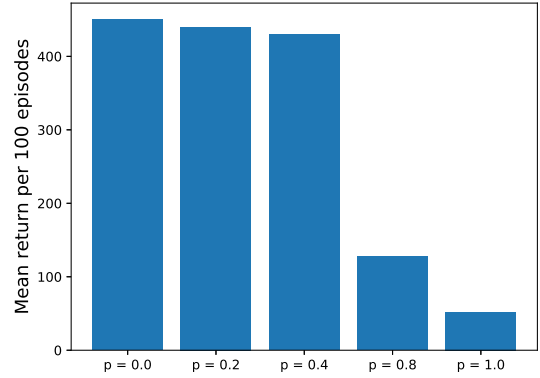}
		\caption{Breakout-$\epsilon$-greedy}
	\end{subfigure}
	\begin{subfigure}[!h]{0.45\textwidth}
	\includegraphics[width=\textwidth]{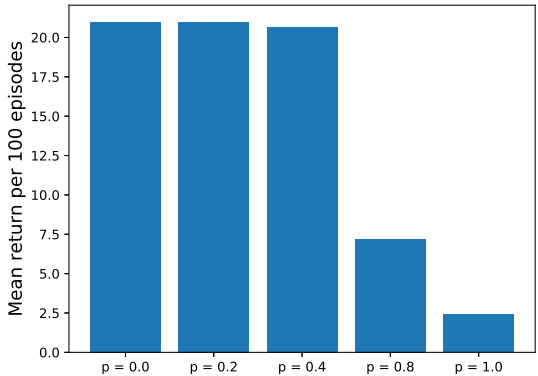}
	\caption{Pong-NoisyNet}
\end{subfigure}
	\begin{subfigure}[!h]{0.45\textwidth}
	\includegraphics[width=\textwidth]{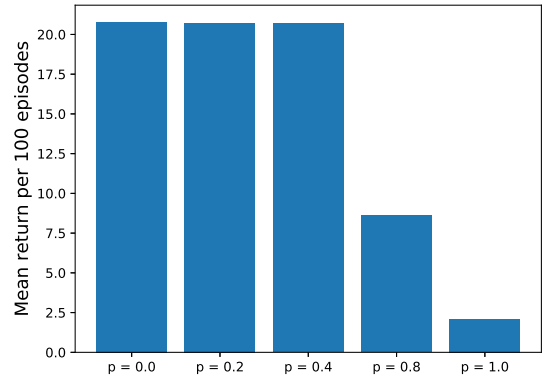}
	\caption{Pong-$\epsilon$-greedy}
\end{subfigure}
	\caption{Test-time performance of adversarially trained policies in non-adversarial conditions ($p = P(attack)$)} 
	\label{test-noatt}
\end{figure*}

Figure \ref{test-noatt} depicts the performance of policies trained under adversarial perturbations at non-adversarial test-time conditions. In concordance with \cite{behzadan2017} and the training results, It can be seen for $p = 0.8$ and $p = 1.0$, test-time performance in non-adversarial conditions severely deteriorates. On the other hand, for $p = 0.2$ and $p = 0.4$, policies trained under adversarial attacks perform almost as well as the original policy in non-adversarial conditions. This result verifies that adversarially trained policies that recover from disruption retain their original performance in non-adversarial conditions. 

\subsection{Test-time Attacks} \label{Results}
\begin{figure}[!h]
	\begin{subfigure}[h]{0.45\textwidth}
		\includegraphics[width=\textwidth]{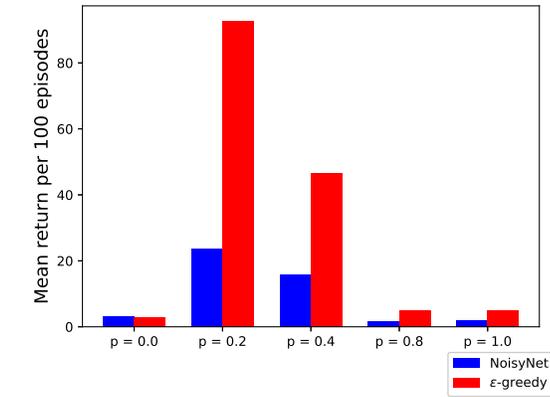}
		\caption{Breakout}
	\end{subfigure}
	\begin{subfigure}[h]{0.45\textwidth}
		\includegraphics[width=\textwidth]{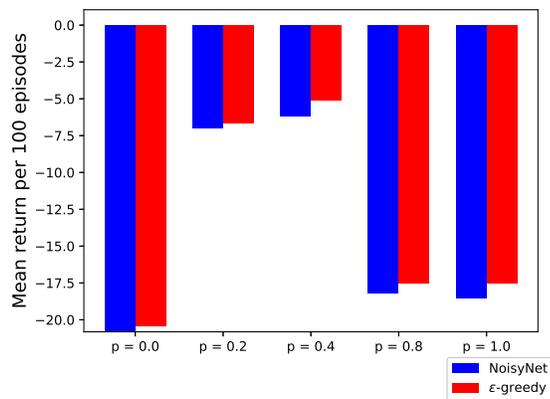}
		\caption{Pong}
	\end{subfigure}
	\caption{Performance under test-time attacks ($p = P(attack)$)} 
	\label{test-time}
\end{figure}

As described in Section \ref{Model}, we investigate test-time attacks under the worst-case scenario of $p = 1.0$ (contiguous attack). Accordingly, we measure the mean return of each agent trained under adversarial perturbations with various values of $p$ when subjected to a contiguous FGSM attack with the same perturbation threshold as that of training-time attacks. Figure \ref{test-time} illustrates the results, demonstrating the strengthened test-time robustness of adversarially trained policies relative to unperturbed policies. Also, Comparing the mean returns of $\epsilon$-greedy and NoisyNet agents, it is observed that in all cases $\epsilon$-greedy performs better than NoisyNet under test-time attacks. This suggests the superior test-time robustness of $\epsilon$-greedy exploration over NoisyNet. An immediate explanation for this behavior may be the greater number of iterations that $\epsilon$-greedy agents take to reach the phase transition point, as observed in Section \ref{TrainingAttacks}. This allows the agent to sample more interactions with the perturbed environment, and thus learn a policy that overcomes a wider variety of perturbed states compared to NoisyNet. Further verification and analysis of this hypothesis remains a priority for future extensions.

\section{Conclusion} \label{Conclusion}
We reported an experimental study of resilience and robustness in Deep Q-Network (DQN) agents under noncontiguous adversarial perturbations. Investigation of training-time attacks demonstrate that if the probability of perturbation for each observation is less than a threshold, DQNs are able to recover from deteriorating training performance and adjust the policy to overcome adversarial perturbations. Corresponding results show the existence of a phase transition point, at which the semi-monotonic deterioration of policy ends and recovery begins. Comparative analysis of our results suggests that $\epsilon$-greedy exploration is more robust to training-time attacks, meaning that it resists the slope of deterioration better than NoisyNet and the corresponding policy experiences less severe drops in performance. On the other hand, NoisyNet may provide better resilience to such attacks by going through phase transition towards recovery in fewer number of iterations. We also demonstrate that recovered policies perform almost as well as the unperturbed policy in non-adversarial test-time experiments. Furthermore, our investigation of test-time attacks on adversarially trained policies demonstrate that such policies are more robust to test-time perturbations than unperturbed policies. The results also suggest that $\epsilon$-greedy exploration provides stronger test-time robustness than NoisyNet.

The experimental observations of this work present a number of promising venues for further research into characteristics of secure deep RL systems. Analysis of the phase transition point and its relationship with attack probability may provide new insights into the limits of reliable training and facilitate secure design practices. Similarly, deeper investigation of exploration mechanisms and their effect on resilience and robustness of deep RL can lead to the development of enhanced exploration techniques and design guidelines for reliable deep RL systems. We believe that research into such venues is crucial for understanding and developing secure RL, as well as emerging and future RL-based approaches to artificial general intelligence.

\section*{Acknowledgments}

The authors would like to thank Dr. William Hsu for his comments and suggestions on this work.

\bibliographystyle{named}
\bibliography{vahid}

\end{document}